\documentclass[utf8]{frontiersSCNS} 

\usepackage{url,hyperref,lineno,microtype,subcaption}
\usepackage[onehalfspacing]{setspace}

\usepackage{amsmath,amsfonts}
\usepackage{algorithm}
\usepackage{algorithmic}
\usepackage{graphicx}
\usepackage{textcomp}
\usepackage{xcolor}
\usepackage{booktabs}
\usepackage{multirow}
\usepackage[super]{nth}
\usepackage{tikz}

\newcommand*\circled[1]{\tikz[baseline=(char.base)]{
		\node[shape=circle,draw,inner sep=0.2pt] (char){#1};}}
\newcommand*\circledB[1]{\tikz[baseline=(char.base)]{
            \node[shape=circle,fill,inner sep=0.2pt] (char) {\textcolor{white}{#1}};}}
\usepackage{soul}
\usepackage{setspace}

\usepackage{fancyhdr}
\fancypagestyle{firstpage}
{
  \fancyhf{}
  \chead{\small Accepted for publication at Frontiers in Neuroscience - Section Neuromorphic Engineering.}
  \cfoot{\thepage}
}

\def\keyFont{\fontsize{8}{11}\helveticabold }
\def\firstAuthorLast{Rachmad Vidya Wicaksana Putra {et~al.}} 

\def\Authors{Rachmad Vidya Wicaksana Putra\,$^{1,*}$, Muhammad Abdullah Hanif\,$^{2}$ and Muhammad Shafique\,$^{2}$}

\def\Address{$^{1}$ Embedded Computing Systems, Institute of Computer Engineering, Technische Universit\"at Wien (TU Wien), Vienna, Austria \\
$^{2}$ eBrain Lab, Division of Engineering, New York University Abu Dhabi (NYUAD), Abu Dhabi, United Arab Emirates  
}

\def\corrAuthor{Rachmad Vidya Wicaksana Putra}
\def\corrEmail{rachmad.putra@tuwien.ac.at}

\begin{document}
\onecolumn
\firstpage{1}

\title[RescueSNN: Reliable SNNs under Permanent Faults]{RescueSNN: Enabling Reliable Executions on Spiking Neural Network Accelerators under Permanent Faults} 

\author[\firstAuthorLast]{\Authors} %This field will be automatically populated
\address{} %This field will be automatically populated
\correspondance{} %This field will be automatically populated

\extraAuth{}% If there are more than 1 corresponding author, comment this line and uncomment the next one.
%\extraAuth{corresponding Author2 \\ Laboratory X2, Institute X2, Department X2, Organization X2, Street X2, City X2 , State XX2 (only USA, Canada and Australia), Zip Code2, X2 Country X2, email2@uni2.edu}

\maketitle
\thispagestyle{firstpage}

%%%%%%%%%%%%%%%%%%%%%%%%%%%%%%%%%%%%%%%%%%%%%%%%%%%%%%%
%%%%%%%%%%%%%%%%%%%%%%%%%%%%%%%%%%%%%%%%%%%%%%%%%%%%%%%

\begin{abstract}
To maximize the performance and energy efficiency of Spiking Neural Network (SNN) processing on resource-constrained embedded systems, specialized hardware accelerators/chips are employed.
However, these SNN chips may suffer from permanent faults which can affect the functionality of weight memory and neuron behavior, thereby causing potentially significant accuracy degradation and system malfunctioning. 
Such permanent faults may come from manufacturing defects during the fabrication process, and/or from device/transistor damages (e.g., due to wear out) during the run-time operation. 
However, the impact of permanent faults in SNN chips and the respective mitigation techniques have not been thoroughly investigated yet. 
Toward this, we propose RescueSNN, a novel methodology to mitigate permanent faults in the compute engine of SNN chips \textit{without requiring additional} retraining, thereby significantly cutting down the design time and retraining costs, while maintaining the throughput and quality. 
The key ideas of our RescueSNN methodology are (1) analyzing the characteristics of SNN under permanent faults; (2) leveraging this analysis to improve the SNN fault-tolerance through effective fault-aware mapping (FAM); and (3) devising lightweight hardware enhancements to support FAM.
Our FAM technique leverages the fault map of SNN compute engine for (i) minimizing weight corruption when mapping weight bits on the faulty memory cells, and (ii) selectively employing faulty neurons that do not cause significant accuracy degradation to maintain accuracy and throughput, while considering the SNN operations and processing dataflow.  
The experimental results show that our RescueSNN improves accuracy by up to 80\% while maintaining the throughput reduction below 25\% in high fault rate (e.g., 0.5 of the potential fault locations), as compared to running SNNs on the faulty chip without mitigation.
In this manner, the embedded systems that employ RescueSNN-enhanced chips can efficiently ensure reliable executions against permanent faults during their operational lifetime.

\tiny
\keyFont{\section{Keywords:} spiking neural networks, accelerators, fault tolerance, manufacturing defects, reliability, resilience, permanent faults.} 
\end{abstract}

%%%%%%%%%%%%%%%%%%%%%%%%%%%%%%%%%%%%%%%%%%%%%%%%%%%%%%%
%%%%%%%%%%%%%%%%%%%%%%%%%%%%%%%%%%%%%%%%%%%%%%%%%%%%%%%
\section{Introduction}
\label{Sec_Intro}

\begin{figure}[t]
\centering
\includegraphics[width=\linewidth]{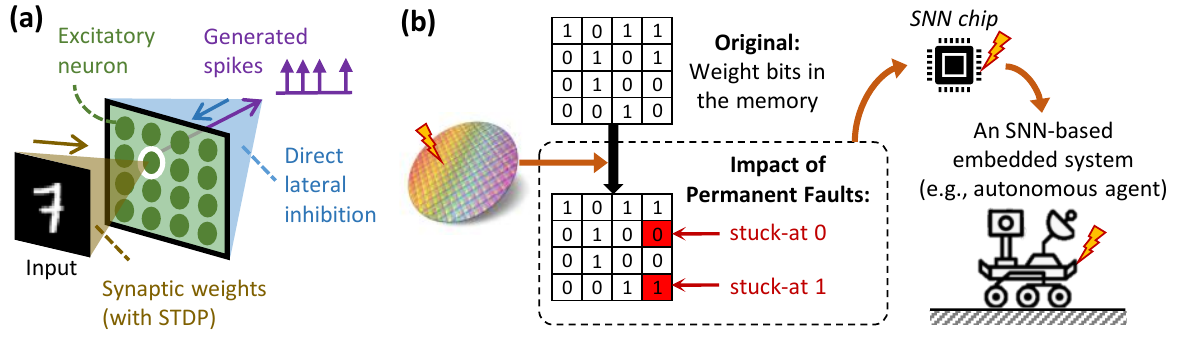}
\caption{(a) An SNN architecture that achieves high accuracy in unsupervised learning scenarios, i.e., a single-layer fully-connected (FC) network~\citep{Ref_Putra_FSpiNN_TCAD20}. (b) Permanent faults in the weight memory of the SNN compute engine may exist in form of stuck-at 0 and stuck-at 1 faults.}
\label{Fig_SNN_PermanentFaults}
\end{figure}

In recent years, SNNs have shown a potential for achieving high accuracy with ultra-low power/energy consumption due to their sparse spike-based operations~\citep{Ref_Putra_FSpiNN_TCAD20}.
Moreover, SNNs can perform unsupervised learning with unlabeled data using spike-timing-dependent plasticity (STDP), which is highly desired for real-world applications (e.g., autonomous agents like UAVs and robotics), especially due to two reasons: these systems are typically subjected to unforeseen scenarios~\citep{Ref_Putra_SpikeDyn_DAC21, Ref_Putra_lpSpikeCon_IJCNN22}; and gathering unlabeled data is cheaper than labeled ones~\citep{Ref_Rathi_PruneQuantizeSNN_TCAD18}. 
An SNN architecture supporting unsupervised learning is shown in Fig.~\ref{Fig_SNN_PermanentFaults}(a).
To maximize the performance and energy efficiency of SNN processing, specialized SNN accelerators/chips are employed~\citep{Ref_Painkras_SpiNNaker_JSCC13, Ref_Akopyan_TrueNorth_TCAD15, Ref_Davies_Loihi_MM18, Ref_Frenkel_ODIN_TBCAS19}.
However, these SNN chips may suffer from permanent faults, which can occur during: (1) \textit{chip fabrication process} due to manufacturing defects, as fabricating an SNN chip with millions-to-billions of nano-scale transistors with 100\% correct functionality is difficult, and even worsen due to the aggressive technology scaling~\citep{Ref_Zhang_PermanentFaults_VTS18, Ref_Hanif_RobustML_IOLTS18, Ref_Hanif_RobustML_Springer21}; and (2) \textit{run time operation} due to device/transistor wear out and damages, that are caused by Hot Carrier Injection (HCI), Bias Temperature Instability (BTI), electromigration, or Time Dependent Dielectric Breakdown (TDDB)~\citep{Ref_Radetzki_FaultToleranceNoC_CSUR13, Ref_Werner_PermanentFaults_CSUR16, Ref_Hanif_RobustML_IOLTS18, Ref_Baloch_Defender_Access19, Ref_Mercier_PermanentFaults_ICCD20, Ref_Hanif_RobustML_Springer21}.

\begin{figure}[t]
\centering
\includegraphics[width=0.82\linewidth]{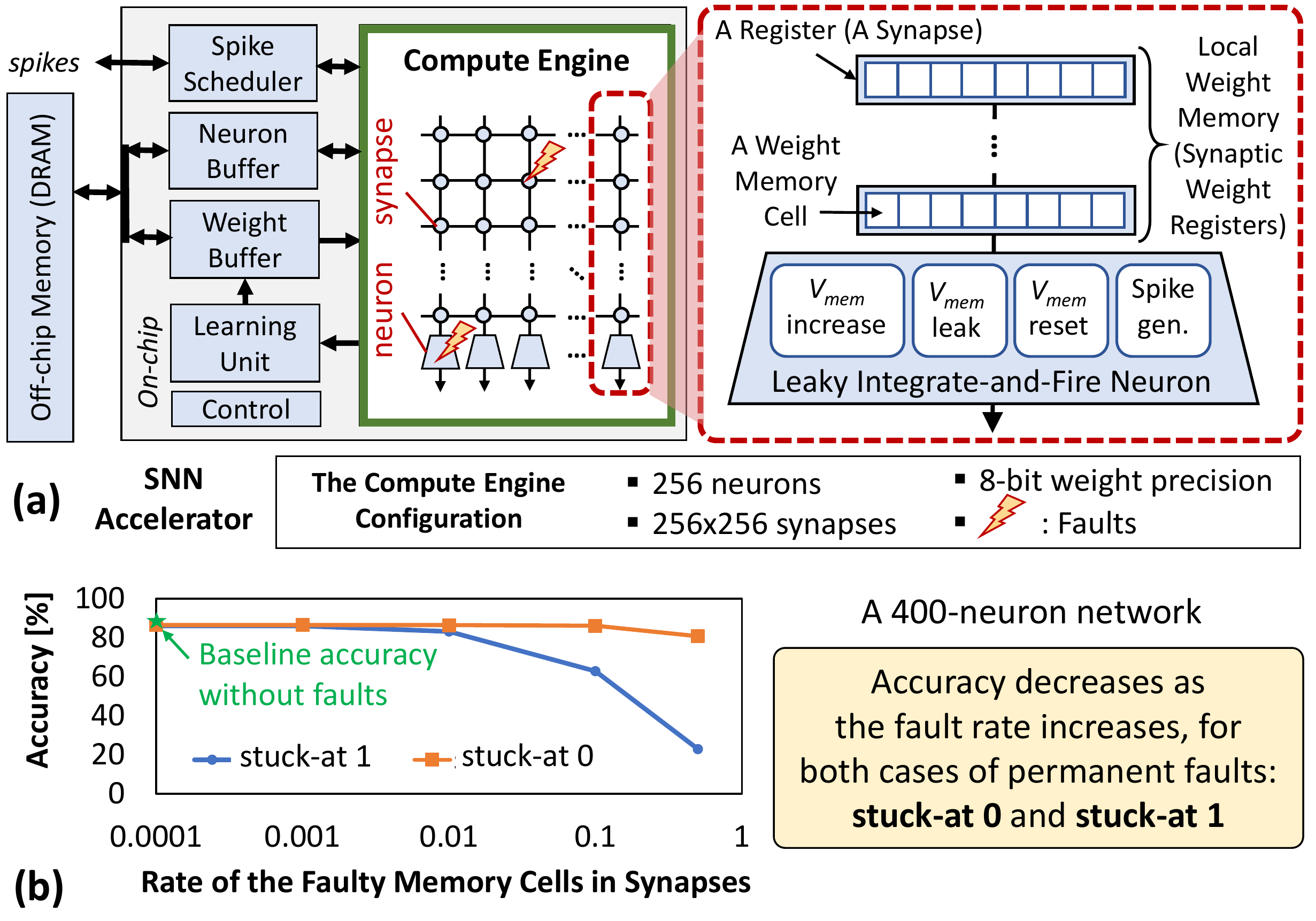}
\caption{(a) The typical SNN accelerator architecture employs crossbar-based synaptic connections~\citep{Ref_Basu_SNNchips_CCIC22}. Synapses and neurons can be affected by permanent faults, whose detailed discussion is provided in Section~\ref{Sec_Backgrounds_FaultModel}. (b) The stuck-at faults in the local weight memory (synaptic weight registers) can decrease accuracy.}
\label{Fig_Observe_PermanentFaults}
\vspace{-0.4cm}
\end{figure}

Permanent faults can affect the functionality of the compute engine of SNN accelerators/chips, including the local weight memory/registers and neurons, by corrupting the weight values and neuron behavior (i.e., membrane potential dynamics and spike generation), thereby degrading the accuracy, as shown in Fig.~\ref{Fig_Observe_PermanentFaults}. 
For instance, permanent faults can change the weight values through stuck-at 0 and 1, as shown in Fig.~\ref{Fig_SNN_PermanentFaults}(b). 
Simply stopping the executions on faulty chips at run time will lead to a short operational lifetime, while discarding the faulty chips at design time will lead to low yield and increase the per-unit cost of the non-faulty chip. 
Therefore, \textit{alternate low-cost solutions for mitigating permanent faults in the SNN compute engine\footnote{For conciseness, we use ``SNN compute engine" or ``compute engine" interchangeably to represent the compute engine of an SNN accelerator/chip.} are required}. 
These solutions will prolong the operational lifetime of SNN chips.
Moreover, such solutions also increase the applicability of wafer-scale chips for SNNs where embracing permanent faults is important to maintain the yield.
    
\textit{\textbf{Targeted Problem:} 
How can we efficiently mitigate permanent faults in the SNN compute engine (i.e., the local weight registers and neurons) on the accuracy, thereby improving the SNN fault tolerance and maintaining the throughput.}
The efficient solution to this problem will enable reliable SNN executions on faulty chips without the need for retraining for energy-constrained embedded systems, such as IoT-Edge devices and autonomous agents. 

%%%%%%%%%%%%%%%%%%%%%%
\subsection{State-of-the-Art and Their Limitations}
\label{Sec_Intro_SOA}

Besides discarding the faulty chips, the standard VLSI fault tolerance techniques like Dual Modular Redundancy (DMR)~\citep{Ref_Vadlamani_DMR_DATE10}, Triple Modular Redundancy (TMR)~\citep{Ref_Lyons_TMR_IBM62}, and Error Correction Code (ECC)~\citep{Ref_Sze_ECCs_USPatent00}, may be used for mitigating permanent faults. 
However, they require extra (redundant) hardware and/or executions which incur huge area and energy overheads. 
State-of-the-art works have studied different design aspects for understanding faults in SNNs and devising mitigation techniques, as follows. 
    
\begin{itemize}
    \item \textbf{SNN fault modeling:} 
    Possible faults that can affect an SNN have been identified in~\citet{Ref_Vatajelu_ReliabilitySNN_VTS19}. 
    In the analog domain, fault modeling for analog neuron hardware and its fault tolerance strategies have been investigated in~\citet{Ref_Sayed_NeuronFaultModel_IOLTS20} and ~\citet{Ref_Spyrou_NeuronFT_DATE21}, which are out of the scope of this work as we target SNN implementation in the digital domain. 
    \item \textbf{SNN fault tolerance:} 
    Previous works studied the impact of faults on SNN weights without considering the underlying hardware architectures and processing dataflows~\citep{Ref_Schuman_ResilinceSNN_IJCNN20, Ref_Rastogi_AstrocytesSTDP_FNINS21} and each discussing a specific fault, like bit-flip or synapse removal.
    Recent works devised mitigation techniques for faults in the weight memories of an SNN hardware~\citep{Ref_Putra_ReSpawn_ICCAD21, Ref_Putra_SparkXD_DAC21, Ref_Putra_EnforceSNN_FNINS22}, while work of \citet{Ref_Putra_SoftSNN_DAC22} aimed at addressing transient faults.
\end{itemize}

In summary, the state-of-the-art works still focus on permanent fault modeling and injection only on the weight memories of SNN hardware. 
Hence, \textit{the impact of permanent faults in the SNN compute engine (i.e., synapses and neurons) and the respective fault mitigation techniques, especially with a focus on avoiding retraining costs, are still unexplored}.
To study the challenges of mitigating permanent faults, we perform an experimental case study in Section~\ref{Sec_Intro_CaseStudy}. 

%%%%%%%%%%%
\subsection{Case Study and Research Challenges}
\label{Sec_Intro_CaseStudy}

In this case study, we consider an SNN accelerator architecture in Fig.~\ref{Fig_Observe_PermanentFaults}(a).
We assume all neurons are not faulty, and inject permanent faults (i.e., stuck-at 0 or 1) on the weight registers with random distribution and different rates of faulty memory cells to see the significance of faulty registers on accuracy. 
Details on the experimental setup are discussed in Section~\ref{Sec_Evaluation}.
From the experimental results in Fig.~\ref{Fig_Observe_PermanentFaults}(b), we make the following key observations.
\begin{itemize}
    \item Classification accuracy decreases as the rate of faulty memory cells increases for both stuck-at 0 and stuck-at 1 scenarios, thereby showing the negative impact of permanent faults in the synapses.  
    \item In the stuck-at 0 case, the stored weight value will either stay the same or decrease from the original value. 
    In the case of decreased weight value, the corresponding neuron will require more stimulus (input spikes) to increase its membrane potential and reach the threshold potential for generating a spike, which represents recognition of a specific class. 
    However, in an SNN model, multiple neurons may be responsible to recognize the same class. 
    Therefore, if the neuron with faulty weight bits cannot recognize the input class, then other neurons may recognize it. 
    As consequence, the accuracy degradation caused by stuck-at 0 in memory cells is relatively small and negligible in some cases.
    \item In the stuck-at 1 case, the stored weight value will either stay the same or increase from the original value. 
    In the case of increased weight value, the corresponding neuron will require less stimulus (input spikes) to increase its membrane potential and reach the threshold potential for generating a spike, which represents recognition of a specific class.
    Therefore, this neuron may become more active to generate more spikes for any input classes, which leads to more misclassification. 
    As consequence, the accuracy degradation caused by stuck-at 1 in memory cells is more significant/noticeable than the stuck-at 0 case.
    \item Combinations of fault types and fault rates lead to different accuracy, which represents different fault patterns in real-world chips, rendering it unpredictable at design time. 
\end{itemize}

Based on these observations, we outline the following research challenges to devise an efficient solution for the targeted problem.
\begin{itemize}
    \item \textit{The mitigation technique should not employ retraining}, as retraining needs huge compute/memory costs, processing time, and a training dataset that may not be available in certain cases due to the restriction policies. 
    Moreover, retraining is not a scalable approach considering the large number of fabricated chips, as it needs to consider a unique fault map from each chip thereby retraining per chip.
    Note, the fault map information can be obtained through the standard wafer/chip test procedure after fabrication, hence this test does not introduce new cost and only incurs a typical cost for chip test~\citep{Ref_Xu_WaferWAT_Access20, Ref_Fan_FaultWaferChip_TASE22}.   
    \item \textit{The mitigation should have minimal performance/energy overhead at run time} as compared to that of the baseline design without fault mitigation technique, thereby making it applicable for energy-constrained embedded systems. 
    \item \textit{The technique should not avoid the use of faulty SNN components (i.e., synapses and neurons)}, as it means omitting the entire computations in the respective columns of the SNN compute engine, which leads to throughput reduction.
    \item \textit{SNNs require a specialized permanent fault mitigation technique} as compared to other neural network computation models (e.g., deep neural networks), since SNNs have different operations and dataflows.
\end{itemize}

%%%%%%%%%%%
\subsection{Our Novel Contributions}
\label{Sec_Intro_NovelContrib}

\begin{figure}[t]
\centering
\includegraphics[width=0.88\linewidth]{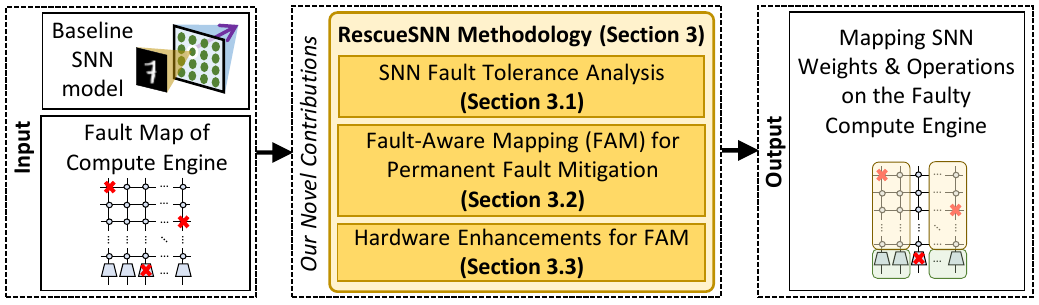}
\caption{Overview of our novel contributions.}
\label{Fig_RescueSNN_Novelty}
\end{figure}

To address the above challenges, we propose RescueSNN, \textit{a novel methodology that enables reliable executions on SNN accelerators under permanent faults}. 
To the best of our knowledge, this work is the first effort that
mitigates permanent faults in the SNN accelerators/chips. 
Following are the key steps of the RescueSNN methodology (the overview is shown in Fig.~\ref{Fig_RescueSNN_Novelty}).
\begin{itemize}
    \item \textbf{Analyzing the SNN fault tolerance} to understand the impact of faulty components (i.e., synapses and neurons) on accuracy considering the given fault rates. 
    \item \textbf{Employing the fault-aware mapping (FAM) techniques} to safely map SNN weights and neuron operations on faulty compute engine, thereby maintaining accuracy and throughput. 
    Our FAM techniques leverage the fault map of the compute engine to perform the following key mechanisms.
    \begin{enumerate}
        \item Mapping the significant weight bits on the non-faulty memory cells of the synapses (weight registers) to minimally pollute/change the weight values.
        \item Selectively employing faulty neurons that do not cause significant accuracy degradation at inference, based on the behavior of their membrane potential dynamics and spike generation.
    \end{enumerate}
    \item \textbf{Devising simple hardware enhancements} to enable efficient FAM techniques. 
    Our enhancements shuffle the weight bits from the synapses by employing simple combinational logic units, such as multiplexers, so that these weight bits can be used for SNN computations.
\end{itemize}

\textbf{Key Results:} 
We evaluate our RescueSNN methodology using Python-based simulations~\citep{Ref_Hazan_BindsNET_FNINF18} on a multi-GPU machine. 
The experimental results show that our RescueSNN improves the SNN accuracy without retraining by up to 80\% and 47\% on the MNIST and the Fashion MNIST respectively, from the SNN processing without fault mitigation. 

%%%%%%%%%%%%%%%%%%%%%%%%
%%%%%%%%%%%%%%%%%%%%%%%%
\section{Backgrounds}
\label{Sec_Backgrounds}

%%%%%%%%%%%
\subsection{Spiking Neural Networks (SNNs)}
\label{Sec_Backgrounds_SNNs}

\textbf{Overview:} 
SNNs are the brain-inspired computational models that employ action potentials (spikes) to encode the information~\citep{Ref_Maass_SNN_NeuNet97, Ref_Putra_Mantis_arXiv22}. 
In an SNN model, neurons and synapses are connected in a specific architecture~\citep{Ref_Tavanaei_DLSNN_Neunet18, Ref_Mozafari_SpykeTorch_FNINS19}.  
In this work, we consider a fully-connected (FC)-based network architecture as shown in Fig.~\ref{Fig_SNN_PermanentFaults}(a), since it has demonstrated a high accuracy when employing bio-plausible learning rules, e.g., the spike-timing-dependent plasticity (STDP). 
This network architecture connects each data of the input (e.g., a pixel of an image) to all excitatory neurons for generating spikes. 
Each spike is used for inhibiting the excitatory neurons except the one from which the spike is generated.
Such bio-plausible learning rules offer efficient learning mechanisms as they perform local learning in each synapse by leveraging spike events without any global loss function, thereby enabling unsupervised and online learning capabilities, which are especially beneficial for autonomous agents~\citep{Ref_Putra_lpSpikeCon_IJCNN22, Ref_Pfeiffer_DLSNN_FNINS18}. 
In this work, we employ the pair-based weight-dependent STDP and bound each weight value ($wgh$) within the defined range, i.e., $wgh = [0,1]$, because this approach has been widely used by previous works~\citep{Ref_Diehl_STDPmnist_FNCOM15, Ref_Srinivasan_EnhPlast_IJCNN17, Ref_Hazan_LMSNN_AMAI19}.
Here, the pair-wise weight-dependent STDP learning rule is used, as it defines the maximum allowed weights, which is suitable for fixed-point format (see Eq.~\ref{Eq_PairWeightSTDP}). 
\begin{equation}
\vspace{-0.3cm}
\small
\begin{split}
\Delta wgh = 
\begin{cases}
-\eta_{pre} \cdot x_{post} \cdot wgh^\mu & \text{on} \; \text{presynaptic spike}\\
\eta_{post} \cdot x_{pre} \cdot (wgh_{m}-wgh)^\mu & \text{on} \; \text{postsynaptic spike}
\end{cases}
\label{Eq_PairWeightSTDP}
\end{split}
\end{equation}
\vspace{0.1cm}
\\
$\Delta wgh$ denotes the weight update, $\eta_{pre}$ and $\eta_{post}$ denote the learning rate for pre- and post-synaptic spike, while $x_{pre}$ and $x_{post}$ denote the pre- and post-synaptic traces, respectively.
$wgh_m$ denotes the maximum allowed weight, $wgh$ denotes the current weight, and $\mu$ denotes the weight dependence factor.
We consider the Leaky Integrate-and-Fire (LIF) neuron model, as it has low computational complexity with high bio-plausible behavior~\citep{Ref_Izhikevich_CompareModels_TNN04}.
Meanwhile, the synapse is modeled by weight value ($wgh$) which represents the strength of the synaptic connection between the corresponding neurons. 
Furthermore, an SNN model typically employs a specific spike coding scheme to encode/decode data information into/from spikes. 
In this work, we consider the rate coding scheme which employs the frequency of spikes to proportionally represent the data, i.e., a higher data value is represented by a higher number of spikes. 
The reason is that, the rate coding scheme achieves high accuracy using bio-plausible STDP-based learning rules, which perform efficient learning mechanisms under unsupervised settings, thereby enabling energy-efficient and smart computing systems~\citep{Ref_Diehl_STDPmnist_FNCOM15, Ref_Rathi_PruneQuantizeSNN_TCAD18, Ref_Putra_FSpiNN_TCAD20}.

\begin{figure}[t]
\centering
\includegraphics[width=0.82\linewidth]{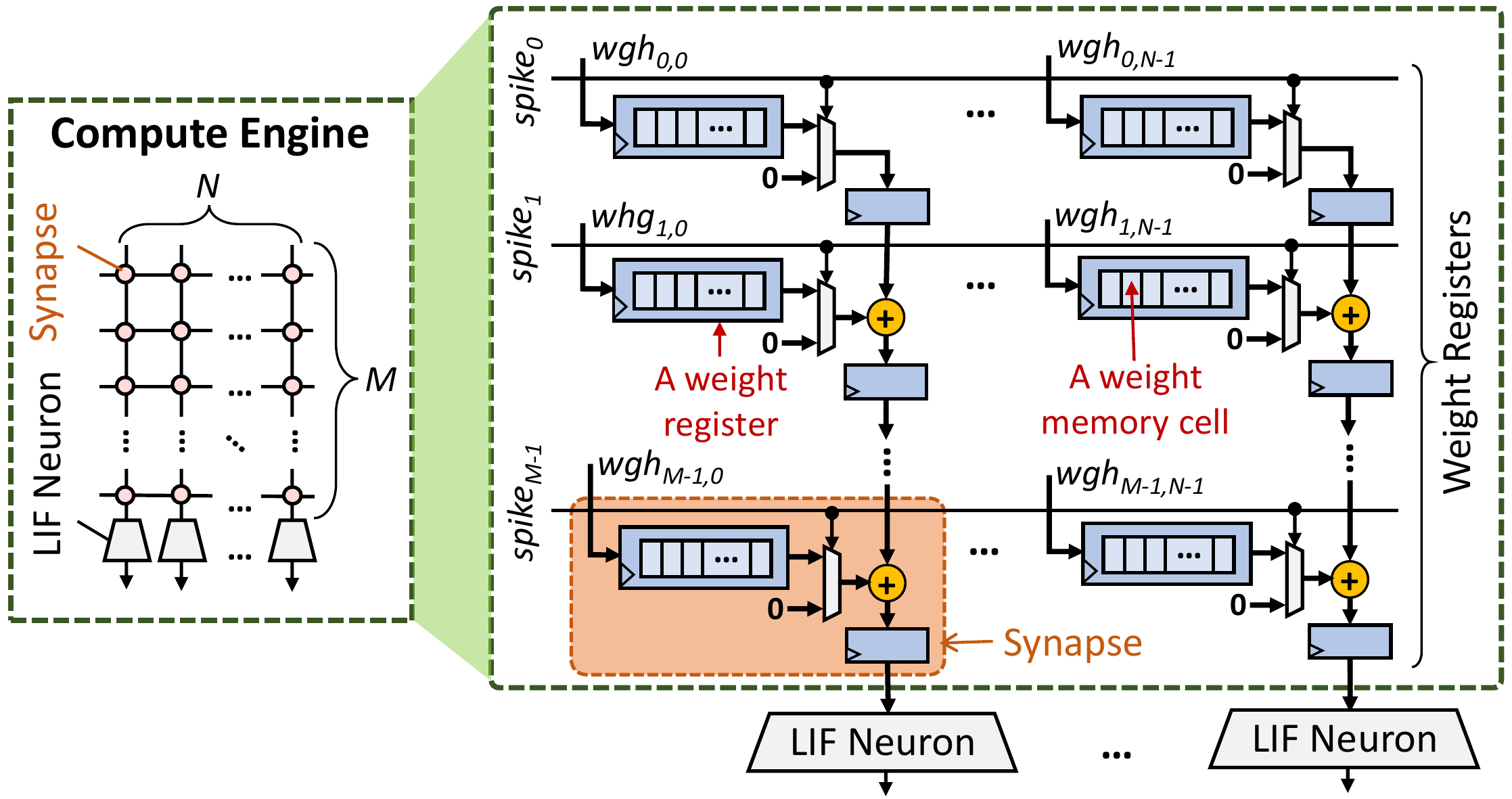}
\caption{The architecture of the compute engine of SNN accelerators.}
\label{Fig_SNNacc_Engine}
\vspace{-0.3cm}
\end{figure}

\smallskip
\textbf{SNN Accelerator Architecture:}
We consider the typical SNN accelerator and compute engine architectures shown in Fig.~\ref{Fig_Observe_PermanentFaults}(a) and Fig.~\ref{Fig_SNNacc_Engine} respectively, which are adapted from the design in \cite{Ref_Frenkel_ODIN_TBCAS19}.
We focus on the compute engine, as it is responsible for generating spikes which determine the accuracy. 
The compute engine has a synapse crossbar, and each synapse has a register that stores a weight value. 
We use 8-bit weight precision as it has a good trade-off for accuracy-memory~\citep{Ref_Putra_FSpiNN_TCAD20, Ref_Putra_QSpiNN_IJCNN21}. 
Each bit is stored in a weight memory cell.
To optimize the chip area, each synapse adds its weight with an accumulated value from the previous synapse in the same column, and stores the cumulative value in a 32-bit register. 
In this manner, only a single connection is required between a neuron and the connected synapses.
Furthermore, each column of the compute engine has a single LIF neuron. 
If a spike comes to a neuron, the membrane potential ($V_{mem}$) increases by the respective weight ($wgh$), otherwise, the $V_{mem}$ decreases/leaks. 
If the $V_{mem}$ reaches the threshold potential ($V_{th}$), a spike is generated, and then $V_{mem}$ goes back to the reset potential ($V_{reset}$). 
Hence, a LIF neuron has four main operations: (1) $V_{mem}$ increase, (2) $V_{mem}$ leak, (3) $V_{mem}$ reset, and (4) spike generation, as shown in Fig.~\ref{Fig_Observe_PermanentFaults}(a).  
Neuron operations (1)-(3) define the membrane potential dynamics.

%%%%%%%%%%%
\subsection{Permanent Fault Modeling}
\label{Sec_Backgrounds_FaultModel}

\subsubsection{Overview}

An SNN compute engine consists of two main components, i.e., synapses and neurons, which have different hardware circuitry. 
Therefore, we need to define a fault model for each component to achieve fast design space exploration.

\begin{figure}[t]
\centering
\includegraphics[width=0.95\linewidth]{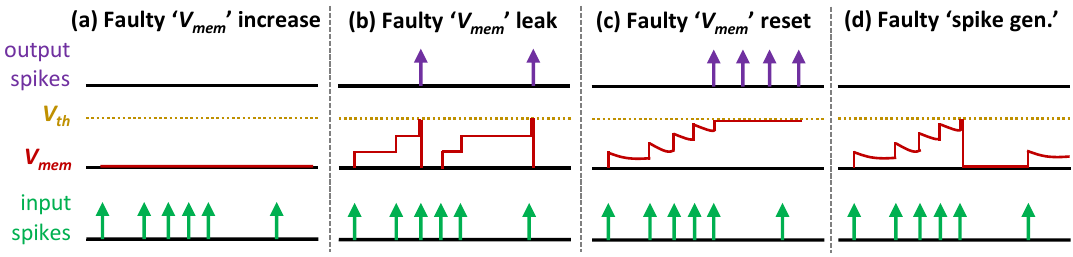}
\vspace{-0.3cm}
\caption{Overview of different faulty LIF neuron operations: (a) faulty `$V_{mem}$ increase', (b) faulty `$V_{mem}$ leak', (c) faulty `$V_{mem}$ reset', and (d) faulty `spike generation'.}
\label{Fig_FaultyNeuronOps}
\end{figure}

\begin{enumerate}
    \item \textbf{Synapses:} 
    Each synapse hardware uses a register to store a weight value.
    Therefore, each permanent fault in a synapse can affect a single weight bit in the form of either a stuck-at 0 or a stuck-at 1 fault. 
    \item \textbf{Neurons:} 
    Each neuron hardware depends on the neuron model to facilitate its operations. 
    Therefore, permanent faults can manifest in different forms depending upon the type of operation being executed on the (digital) neuron hardware, as discussed in the following (see an overview in Fig.~\ref{Fig_FaultyNeuronOps}).
    \begin{itemize}
        \item \textit{Faults in the `$V_{mem}$ increase' operation} make the neuron unable to increase its membrane potential.
        As a result, this neuron cannot generate any spikes (i.e., a dormant neuron).
        \item \textit{Faults in the `$V_{mem}$ leak' operation} make the neuron unable to decrease its membrane potential. 
        Hence, this neuron acts like the Integrate-and-Fire (IF) neuron model.
        \item \textit{Faults in the `$V_{mem}$ reset' operation} make the neuron unable to reset its membrane potential. 
        As a result, this neuron will continuously generate spikes.
        \item \textit{Faults in the `spike generation'} make the neuron unable to generate spikes (i.e., dormant neuron).
    \end{itemize}
\end{enumerate}

%%%%%%%

\begin{figure}[t]
\centering
\includegraphics[width=0.85\linewidth]{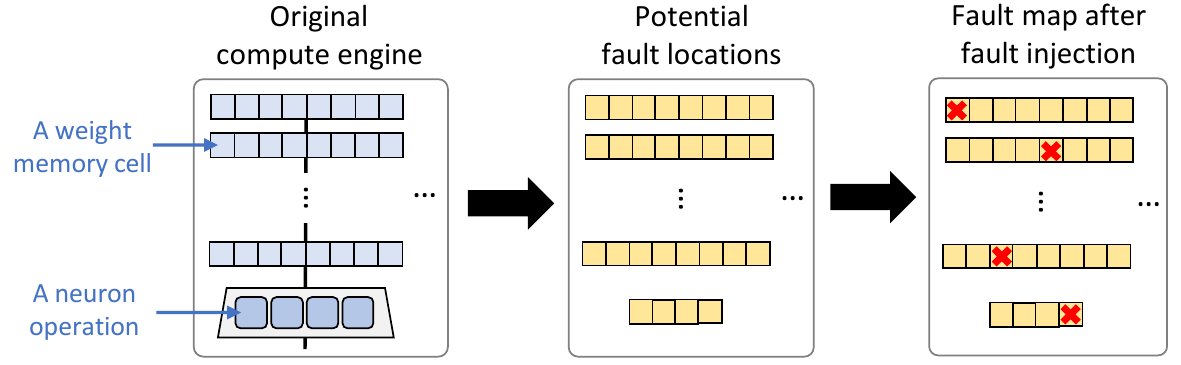}
\vspace{-0.3cm}
\caption{The key steps of permanent fault generation and distribution on the SNN compute engine.}
\label{Fig_FaultMap}
\end{figure}
%%%%%%%

%%%%%%%
\subsubsection{Fault Generation and Distribution}

Previous studies have shown that permanent faults occur in random locations of a chip, hence leading to a certain fault map~\citep{Ref_Zhang_PermanentFaults_VTS18, Ref_Radetzki_FaultToleranceNoC_CSUR13, Ref_Werner_PermanentFaults_CSUR16, Ref_Mercier_PermanentFaults_ICCD20, Ref_Stanisavljevic_NanoCAS_Springer11}. 
Following are the key steps to generate and distribute permanent faults on the SNN compute engine, as shown in Fig.~\ref{Fig_FaultMap}. 
\begin{enumerate}
    \item We consider a weight memory cell and a neuron operation as the potential fault locations. 
    \item We generate permanent faults based on the given fault rate and distribute them randomly across the potential fault locations.  
    The fault rate represents the ratio between the total number of faulty weight memory cells and neuron operations to the total number of potential fault locations (i.e., the total number of weight memory cells and neuron operations).   
    \item If a fault occurs in a local weight memory cell, then we randomly select the type of stuck-at fault (i.e., either stuck-at 0 or stuck-at 1). 
    Meanwhile, if a fault occurs in a neuron operation, then we randomly select the type of permanent faulty operation.
\end{enumerate}

%%%%%%%%%%%%%%%%%%%%%%%%
%%%%%%%%%%%%%%%%%%%%%%%%
\section{RescueSNN Methodology}
\label{Sec_RescueSNN}

\begin{figure}[t]
\centering
\includegraphics[width=0.95\linewidth]{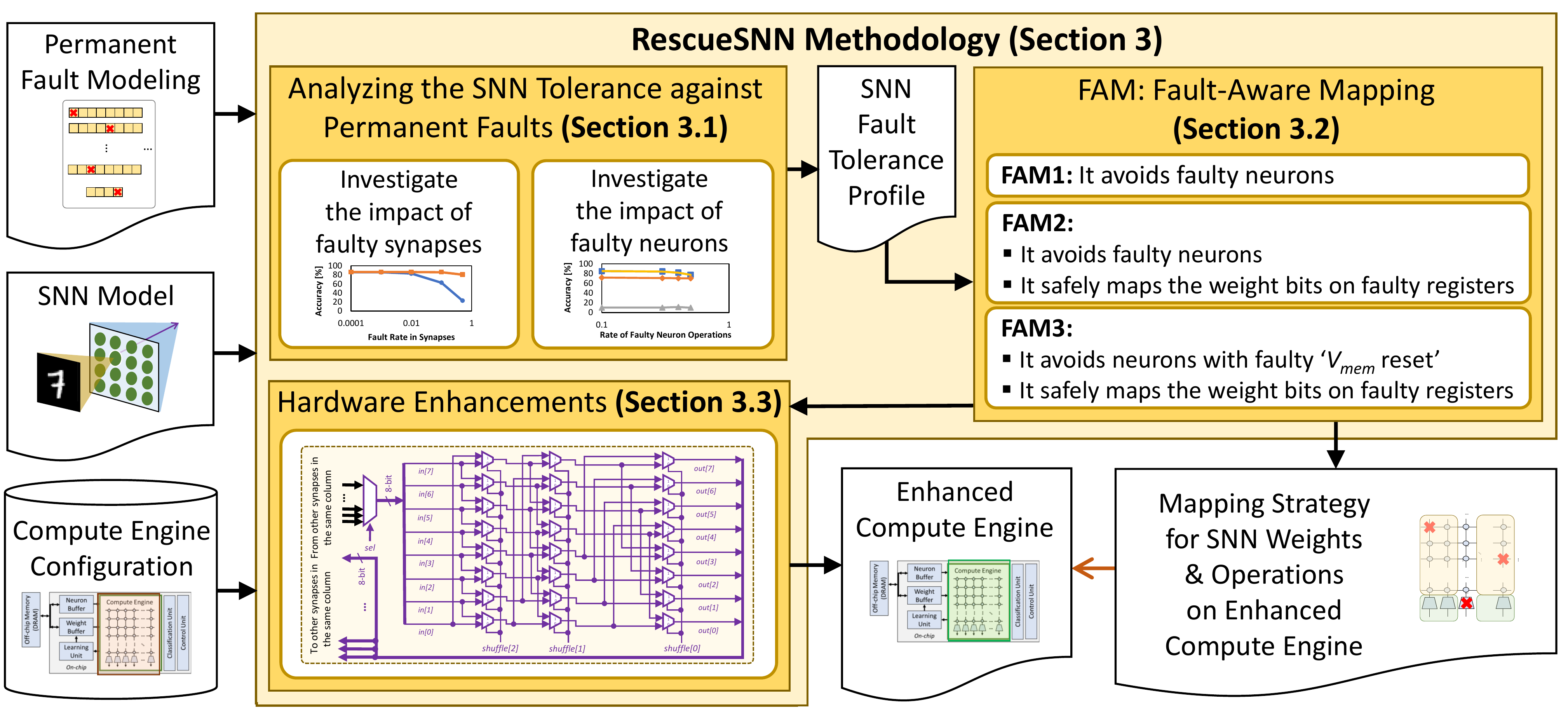}
\vspace{-0.3cm}
\caption{Overview of our RescueSNN methodology and its key steps for mitigating permanent faults in the SNN compute engine.}
\label{Fig_RescueSNN_Overview}
\end{figure}

The overview of our RescueSNN methodology is shown in Fig.~\ref{Fig_RescueSNN_Overview}, and its key steps are discussed in the following subsections.

%%%%%%%%%%%
\subsection{SNN Fault Tolerance Analysis under Permanent Faults}
\label{Sec_RescueSNN_SNN_FTAnalysis}

\begin{figure}[t]
\centering
\includegraphics[width=0.8\linewidth]{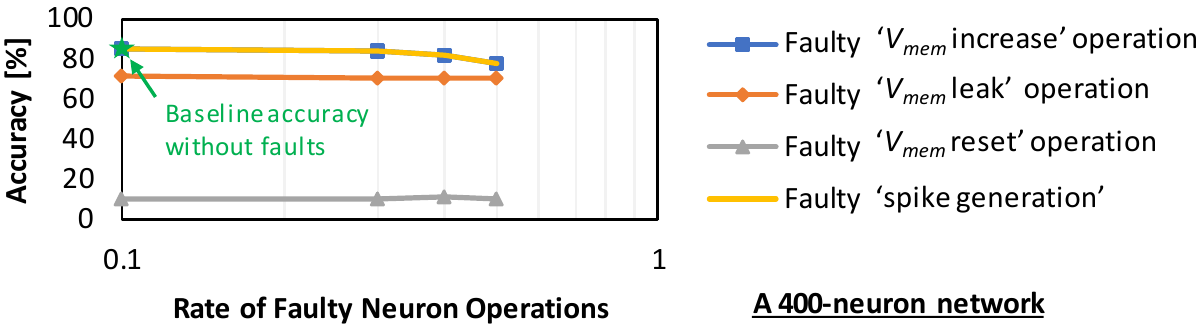}
\caption{Impact of faulty neuron operations on accuracy. Different faulty neuron operations have a different impact on accuracy. Notable accuracy degradation happens when faulty `$V_{mem}$ reset' operations are employed.}
\label{Fig_Observe_FaultyNeuronOps}
\end{figure}

SNN fault tolerance analysis is important to understand how a given SNN model will behave considering a specific operating condition (e.g., a combination of certain fault rate, type of stuck-at fault, architecture of the compute engine, etc.).
This analysis provides information which can be leveraged for devising an efficient fault mitigation technique.
Therefore, \textit{our RescueSNN methodology investigates the interaction between the faulty components (i.e., synapses and neurons) and the obtained accuracy}.
To do this, we perform the following experimental case studies, while considering a 400-neuron network with fully-connected architecture like in Fig.~\ref{Fig_SNN_PermanentFaults}(a). 
\begin{itemize}
    \item \textbf{We study the accuracy under faulty weight registers}, by injecting a specific stuck-at fault (i.e., either stuck-at 0 or stuck-at 1) in the weight registers, while considering fault-free neurons. 
    Experimental results are shown in Fig.~\ref{Fig_Observe_PermanentFaults}(b).
    We observe that both stuck-at 0 and stuck-at 1 faults can degrade accuracy. 
    Therefore, the mitigation technique should address both stuck-at faults.
    \item \textbf{We study the accuracy under faulty neuron operations}, by injecting faults on the neuron operations, while considering fault-free weight registers. 
    Experimental results are shown in Fig.~\ref{Fig_Observe_FaultyNeuronOps}, from which we make the following observations.
    \begin{enumerate}
      \item Faulty `spike generation', `$V_{mem}$ increase', and `$V_{mem}$ leak' operations have tolerable accuracy, since their faulty behavior does not dominate the spiking activity, and/or the function of the corresponding faulty neurons for classification may be substituted by other neurons that recognize the same class. Therefore, these neurons can still be used for SNN processing.
      \item Faulty `$V_{mem}$ reset' operations cause significant accuracy degradation, as these operations make the corresponding neurons dominate classification. 
      Therefore, these neurons should not be used for SNN processing.
    \end{enumerate}
\end{itemize}

Note, complex SNN models with multiple layers and different computational architectures (e.g., convolutional and fully-connected) may have different observation results as compared to results in Fig.~\ref{Fig_Observe_FaultyNeuronOps}. 
However, previous work has observed similar trends to our study, i.e., neurons with faulty `$V_{mem}$ reset' operations continuously generate spikes (so-called \textit{saturated neurons}) and cause the most significant accuracy degradation than other types of faulty neuron operations~\citep{Ref_Spyrou_NeuronFT_DATE21}.
It also identified that, saturated neurons affect classification accuracy at any layer of SNN models, as these faulty neurons always dominate the classification activity which results in a significant accuracy degradation.
This finding is consistent with the insights provided by our study. 
However, it is still challenging to achieve high accuracy when employing STDP-based learning on complex SNN models~\citep{Ref_Rathi_ExploreNeuroComp_CSUR23}, 
thereby hindering their applicability for diverse applications, such as systems with online training and unsupervised learning requirements (e.g., autonomous mobile agents). 
Therefore, in this work, we consider the FC-based SNNs shown in Fig.~\ref{Fig_SNN_PermanentFaults}(a) to enable multiple advantages, such as high accuracy, unsupervised learning capabilities, and efficient online training. 
    
%%%%%%%%%%%
\subsection{The Proposed Fault-Aware Mapping (FAM)}
\label{Sec_RescueSNN_ProposedFAM}

Permanent faults in SNN chips can be identified at the design time and at the run time. 
The post-fabrication test can be employed to find a set of locations of faults (due to manufacturing defects) in the SNN compute engine (i.e., fault map) at the design time~\citep{Ref_Zhang_PermanentFaults_VTS18, Ref_Putra_ReSpawn_ICCAD21}. 
Meanwhile, the online test strategy like the Built-In Self-Test (BIST) techniques can be employed to obtain the fault map (due to device wear out or physical damages) at the run time~\citep{Ref_Baloch_Defender_Access19, Ref_Wang_Design4Test_TC19, Ref_Mercier_BiSuT_TCAD21}. 
\textit{Our RescueSNN methodology leverages this fault map to safely map the SNN weights and operations on the compute engine, thereby minimizing the negative impact of permanent faults}.
To do this, the RescueSNN employs Fault-Aware Mapping (FAM) techniques that mitigate the faults in synapses and neurons through the following key mechanisms.

%%%%
%%%%

\begin{figure}[t]
\centering
\includegraphics[width=0.7\linewidth]{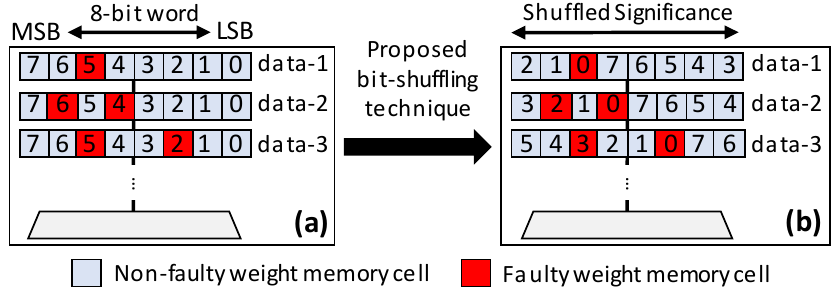}
\caption{(a) Illustration of possible fault locations (fault map) in faulty weight registers. (b) The proposed circular-shift bit-shuffling technique for the corresponding fault map.}
\label{Fig_FaultyWeightRegs}
\end{figure}

\begin{enumerate}
    \item \textbf{FAM for Synapses:} 
    The significant weight bits are placed on the non-faulty weight memory cells, while the insignificant bits are placed on the faulty ones, by performing a simple bit-shuffling technique.
    The significance of weight bits can be identified by experiments that observe the accuracy after modifying a specific bit~\citep{Ref_Putra_ReSpawn_ICCAD21}. 
    In general, previous studies have observed that the significance of a weight bit is proportional to its bit location. 
    For instance, in 8-bit fixed-point precision, bit-7 has the highest significance than other bits.  
    Furthermore, a synapse may have a single faulty bit or multiple faulty bits. 
    Therefore, we propose a mapping strategy that can address both cases using the following steps (see an overview in Fig.~\ref{Fig_FaultyWeightRegs}).
    \begin{itemize}
        \item We identify the faulty weight bits (e.g., through the post-fabrication testing) to obtain information of the fault map and fault rate in each synapse hardware. 
        \item We identify the maximum fault rate in each synapse hardware for safely storing a weight.
        In this work, we consider a maximum of 2 faulty bits from an 8-bit weight, based on the fault rates that offer tolerable accuracy from analysis in Section~\ref{Sec_RescueSNN_SNN_FTAnalysis}.  
        \item We identify the segment in each synapse with the highest number of subsequent non-faulty memory cells. 
        This information is leveraged for maximizing the possibility of storing the significant bits in the non-faulty cells. 
        Hence, we also examine the corner case (i.e., the right-most and left-most cells) as possible subsequent non-faulty memory cells; see the third row of Fig.~\ref{Fig_FaultyWeightRegs}(b) with data-3.
        \item We perform a circular-shift technique for each data word to efficiently implement bit-shuffling.  
    \end{itemize}
    \item \textbf{FAM for Neurons:}
    The use of neurons should be avoided if they have faulty `$V_{mem}$ reset' operations, as these faulty operations cause significant accuracy degradation. 
    Meanwhile, neurons with other types of faults can still be used for SNN processing, as their faulty behavior does not dominate the spiking activity. 
    Different SNN operations that aim at recognizing the same input class are mapped on both the faulty and fault-free neurons for maintaining throughput, while compensating the loss from the faulty `$V_{mem}$ increase', `$V_{mem}$ leak', and `spike generation' operations. 
\end{enumerate}

\begin{figure}[t]
\centering
\includegraphics[width=0.75\linewidth]{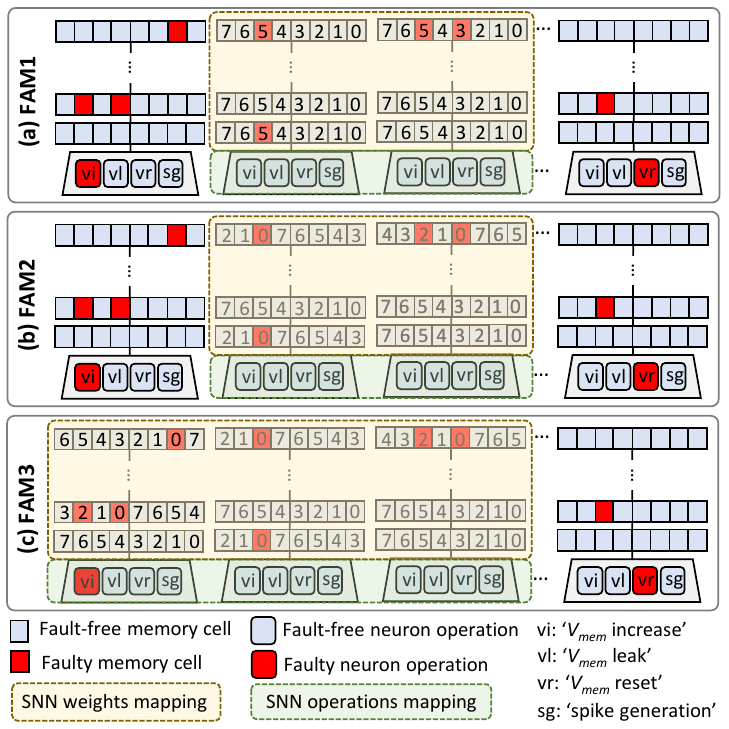}
\caption{Our FAM strategies: (a) FAM1, (b) FAM2, and (c) FAM3.}
\label{Fig_FAMTechniques}
\end{figure}

\smallskip
Afterward, \textit{we leverage these mechanisms for devising three mapping strategies, as the variants of our FAM technique (i.e., FAM1, FAM2, and FAM3)}, which provide trade-offs between accuracy and mapping complexity, as discussed in the following. 
\begin{itemize}
    \item \textbf{FAM1:} 
    It avoids mapping the SNN weights and operations on the columns of compute engine that have faulty neurons, as shown in Fig.~\ref{Fig_FAMTechniques}(a), as faulty neurons can reduce the accuracy more than faulty registers, especially in the case of faulty `$V_{mem}$ reset'.  
    However, FAM1 does not mitigate the negative impact of faults in the registers, hence the accuracy improvement is sub-optimal.
    The benefit of FAM1 is due to its simple mechanism which enables a low-complexity control mechanism.
    \item \textbf{FAM2:} 
    It maps the SNN weights and operations on the columns of compute engine that have fault-free neurons (just like FAM1) and employs a bit-shuffling technique to map the significant weight bits on the non-faulty memory cells, as shown in Fig.~\ref{Fig_FAMTechniques}(b).
    Therefore, FAM2 can improve the SNN fault tolerance at the cost of a more complex control mechanism than FAM1.
    \item \textbf{FAM3:} 
    It selectively maps the SNN weights and operations on the columns of compute engine that do not have faulty `$V_{mem}$ reset' operations, as well as maps the significant weight bits on the non-faulty memory cells using a bit-shuffling technique, as shown in Fig.~\ref{Fig_FAMTechniques}(c).
    Therefore, FAM3 can enhance the SNN fault tolerance as compared to FAM1 at the cost of a more complex control mechanism, and can improve the throughput as compared to FAM1 and FAM2. 
\end{itemize}

Information regarding how to map the SNN weights and operations on the compute engine is provided through software program (e.g., firmware), thereby enabling the applicability and flexibility of the proposed FAM technique (e.g., FAM1, FAM2, or FAM3) for different possible fault maps on the compute engine.
The meta data of this information is stored in the on-chip buffer, which can be accessed for operations in the compute engine.

%%%%%%%%%%%
\subsection{Our Hardware Enhancements for FAM}
\label{Sec_RescueSNN_HWenhance}

%%%

%%%

\begin{figure}[t]
\centering
\includegraphics[width=\linewidth]{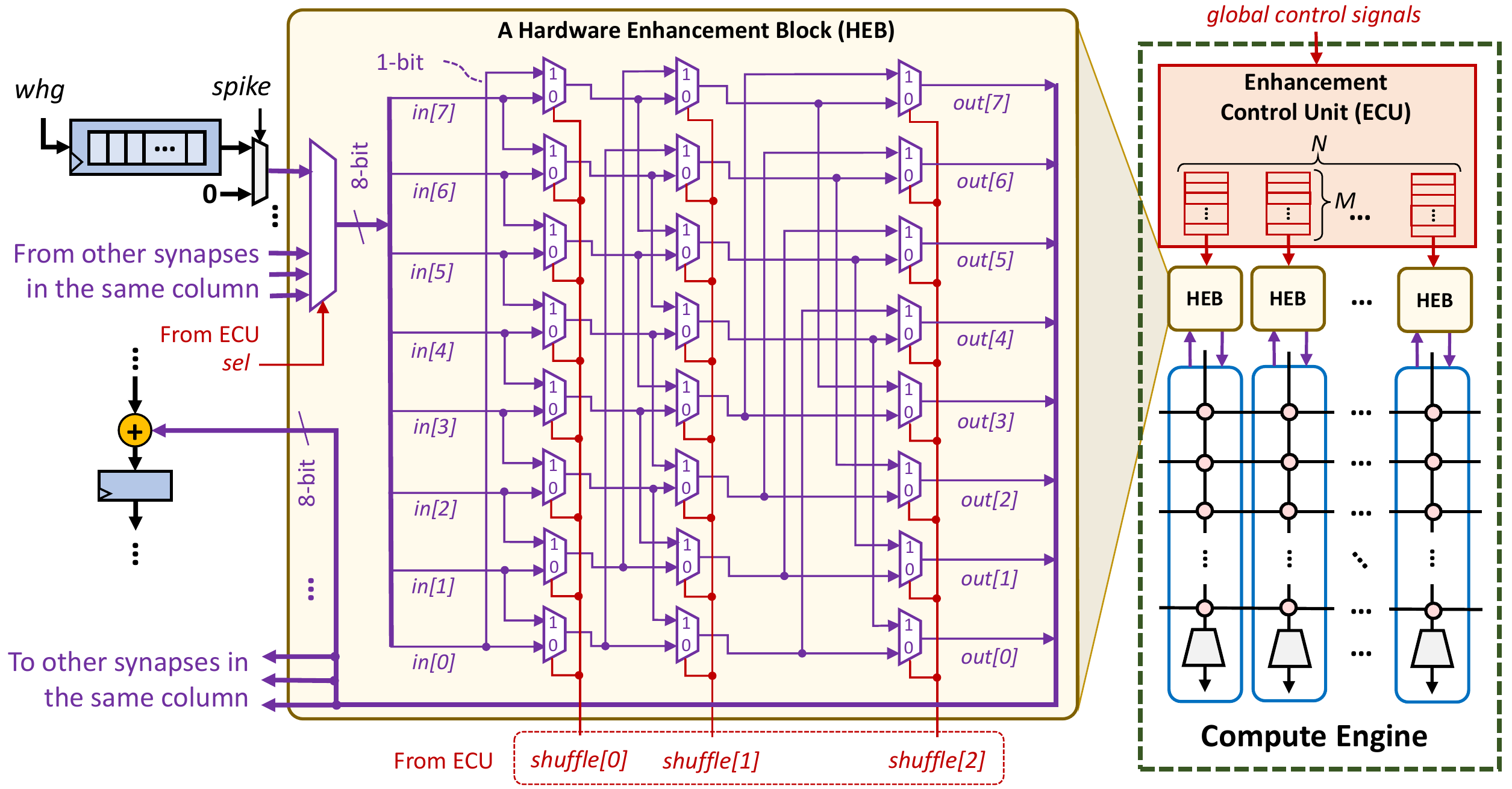}
\caption{The architecture of the proposed enhancements, including the hardware enhancement blocks (HEBs) and the enhancement control unit (ECU), for accommodating FAM strategies.}
\label{Fig_SNNacc_HWenhance}
\end{figure}

Our FAM2 and FAM3 strategies may make the weight bits stored in a shuffled form. 
Therefore, an additional mechanism is required for converting these weight bits into the original order, so that they can be used for SNN executions.
Toward this, \textit{we propose lightweight hardware enhancements to support the re-shuffling mechanism to undo the data transformation, i.e., through a simple 8-bit barrel shifter}. 
The key idea is to re-shuffle the order of output wires from each synapse into the original order, so that the corresponding weights can be used directly for neuron operations. 
To optimize the overheads (e.g., area), we share a \textit{hardware enhancement block} (HEB) with all synapses in the same column of compute engine, and different synapses will access the enhancement block at a different time; see Fig.~\ref{Fig_SNNacc_HWenhance}.
In this manner, the number of HEBs is equal to the number of columns in the SNN compute engine.
Furthermore, to control the functionality of HEBs, we employ an \textit{enhancement control unit} (ECU).
This ECU stores the bit-shifting information and uses it for controlling the barrel shifter in HEBs. 
For each column of the compute engine, the ECU employs (1) a dedicated selector signal \textit{sel} to determine which weight should be processed in the HEB at a time, and (2) a set of registers that stores bit-shifting information \textit{shuffle[2:0]} for all weights in the same column.

%%%%%%%%%%%%%%%%%%%%%%%%
%%%%%%%%%%%%%%%%%%%%%%%%
\section{Evaluation Methodology}
\label{Sec_Evaluation}

\begin{figure}[t]
\centering
\includegraphics[width=0.9\linewidth]{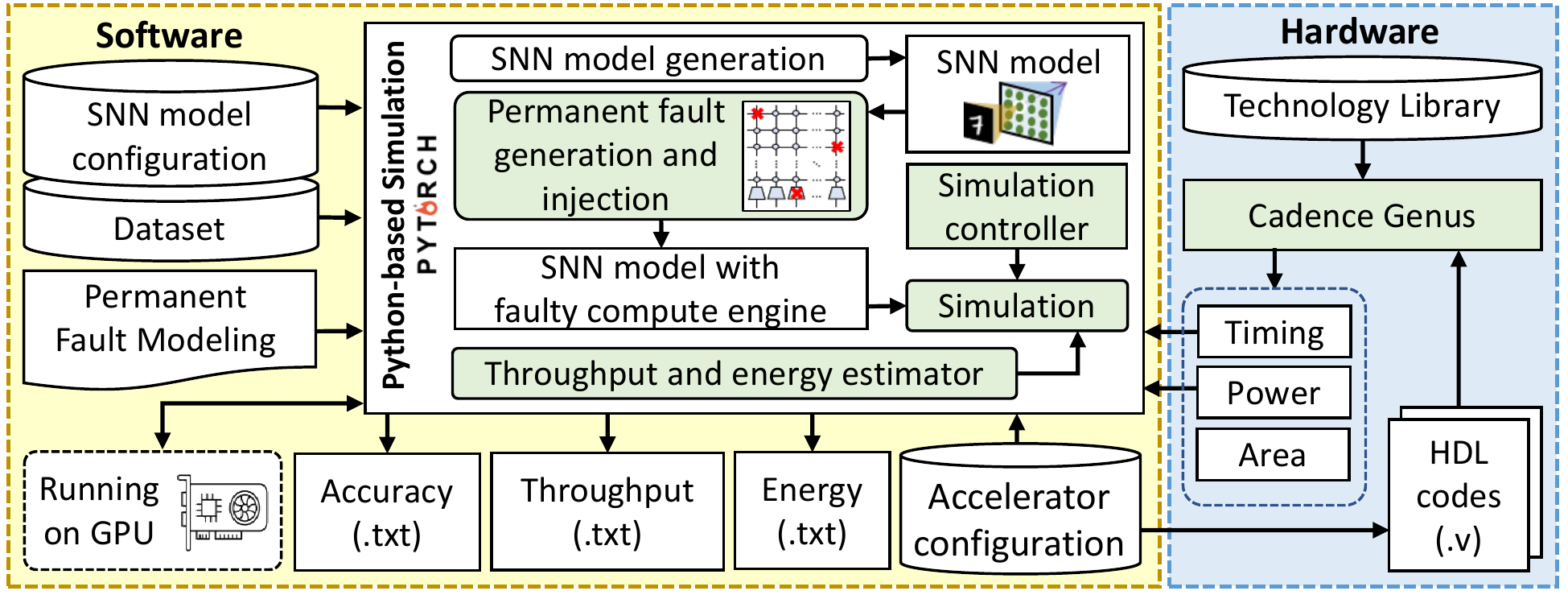}
\caption{Overview of the experimental setup and tools flow.}
\label{Fig_Evaluation_Method}
\end{figure}

For evaluating the RescueSNN methodology, we employ the experimental setup shown in Fig.~\ref{Fig_Evaluation_Method}. 
We use the fully-connected network shown in Fig.~\ref{Fig_SNN_PermanentFaults}(a) with a different number of neurons, to evaluate the generality of our RescueSNN methodology. 
For conciseness, we represent a network with $i$-number of neurons as N$i$. 
We use the MNIST and Fashion MNIST as the workloads, and adopt the same test conditions as used widely by the SNN community~\citep{Ref_Diehl_STDPmnist_FNCOM15}. 
For comparison, we consider the SNN without fault mitigation as the baseline. 

\begin{figure}[t]
\centering
\includegraphics[width=0.75\linewidth]{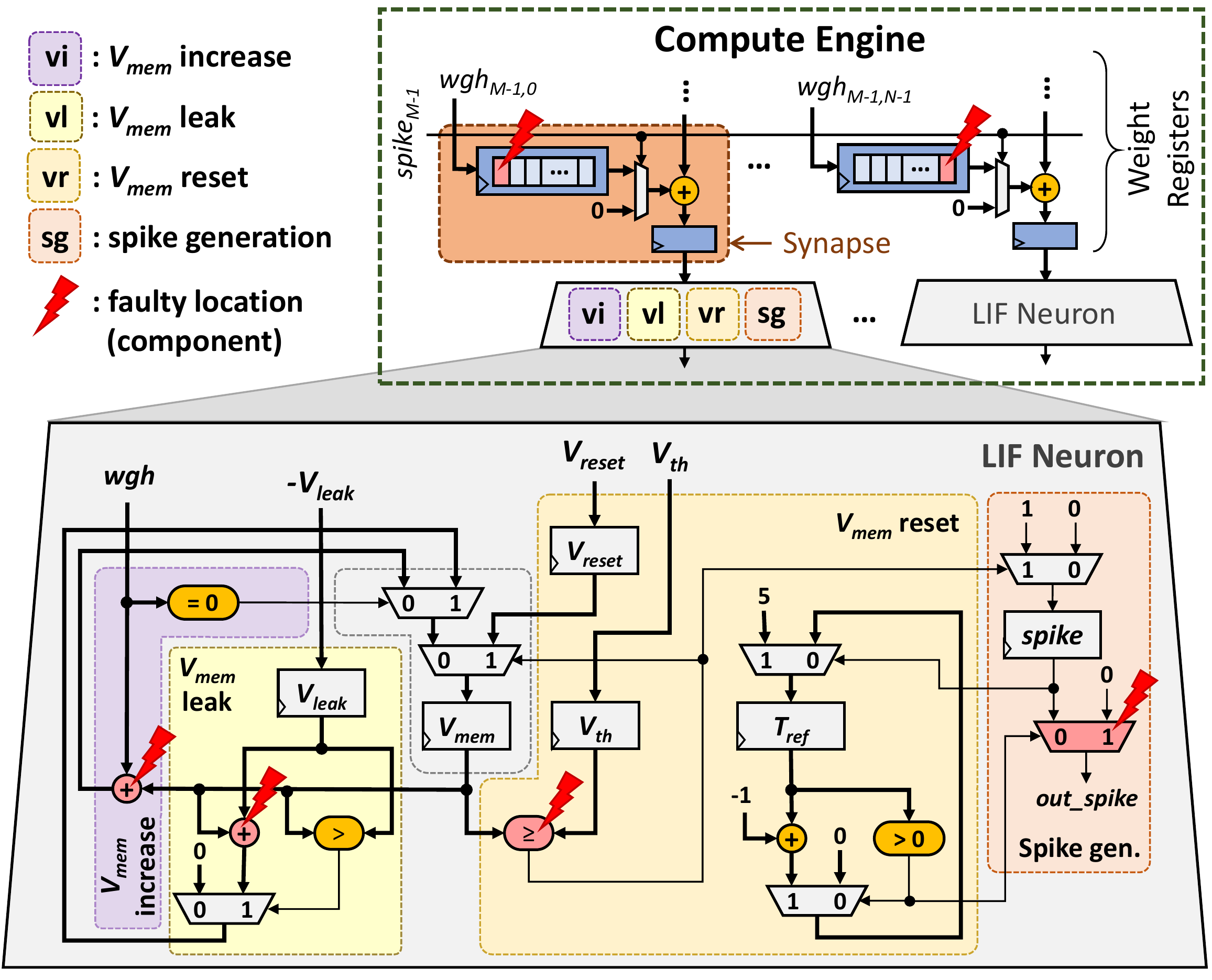}
\caption{The potential locations/components that can be affected by permanent faults to cause faulty memory cells as well as faulty `$V_{mem}$ increase', `$V_{mem}$ leak', `$V_{mem}$ reset', and `spike generation' operations.}
\label{Fig_FaultComponents}
\end{figure}

\textbf{Fault Generation and Injection:}
Permanent faults are generated based on the fault modeling in Section~\ref{Sec_Backgrounds_FaultModel} of the revised manuscript.
To do this, we first generate binary values (i.e., 0 and 1) based on the given fault rate while considering the potential fault locations (shown in Fig.~\ref{Fig_FaultMap}).
Here, `0' represents a non-faulty memory cell in synapses or a non-faulty operation in neurons; while `1' represents a faulty memory cell in synapses or a faulty operation in neurons.
These binary values are then randomly distributed into an array that represents the potential fault locations, so that each value corresponds to a specific weight memory cell or a specific neuron operation.  
Fig.~\ref{Fig_FaultComponents} shows the potential locations/components that can be affected by permanent faults to cause faulty memory cells as well as faulty `$V_{mem}$ increase', `$V_{mem}$ leak', `$V_{mem}$ reset', and `spike generation' operations.
For \textit{each fault in the weight memory cells (synapses)}, we randomly determine the type of fault, i.e., either stuck-at~0 or stuck-at~1. 
In stuck-at~0 case, value 0 is injected to the corresponding memory cell; while in stuck-at 1 case, value 1 is injected.
Meanwhile, \textit{each fault in neurons} corresponds to either faulty `$V_{mem}$ increase', `$V_{mem}$ leak', `$V_{mem}$ reset', or `spike generation' operation (as shown in Fig.~\ref{Fig_FaultyNeuronOps}). 
Each faulty behavior in the corresponding neuron is realized through different approaches as described in the following.
\vspace{-0.2cm}
\begin{itemize}
    \item \textit{Faulty `$V_{mem}$ increase' operation}: 
    It is mainly caused by faulty addition in the `$V_{mem}$ increase' part, hence $V_{mem}$ is not increased despite there are incoming spikes.
    \item \textit{Faulty `$V_{mem}$ leak' operation}: 
    It is mainly caused by faulty subtraction in `$V_{mem}$ leak' part, hence $V_{mem}$ is not decreased despite there are no incoming spikes.
    \item \textit{Faulty `$V_{mem}$ reset' operation}: 
    It is mainly caused by faulty comparison in `$V_{mem}$ reset' part, hence the spike generator is activated to continuously generate spikes.
    \item \textit{Faulty `spike generation'}: 
    It is mainly caused by faulty multiplexing in `spike generation' part, hence the spike generator is always deactivated and no output spikes are produced.
\end{itemize}

\textbf{Accuracy Evaluation:}
We use the Python-based simulations~\citep{Ref_Hazan_BindsNET_FNINF18}, which run on Nvidia RTX 2080 Ti GPUs, while considering SNN accelerator architecture shown in Fig.~\ref{Fig_Observe_PermanentFaults}(a) and \ref{Fig_SNNacc_Engine}.

\textbf{Hardware Evaluation:}
We evaluate the area, energy consumption, and throughput of both the original compute engine (without enhancements) and the enhanced compute engine using our RescueSNN methodology. 
To do this, we design RTL codes for both (original and enhanced) compute engines, then synthesize them using the Cadence Genus tool considering a 65nm CMOS technology to obtain their area, power consumption, and timing (i.e., a clock cycle latency for SNN processing on the compute engine).
Afterward, we calculate the required number of cycles and computation latency for processing an input sample (i.e., \textit{latency-per-sample}), considering the timing from synthesis and the mapping strategy on active synapses and neurons in the compute engine.
Then, we estimate the throughput by computing the number of samples that can be processed within one second of SNN inference based on the information of latency-per-sample.
Furthermore, we also estimate the energy consumption by leveraging the information of power consumption from synthesis and latency-per-sample for SNN inference. 
The estimation of throughput and energy consumption is also performed using the Python-based simulation framework~\citep{Ref_Hazan_BindsNET_FNINF18}.

%%%% insert
\begin{figure*}[t]
\centering
\includegraphics[width=\linewidth]{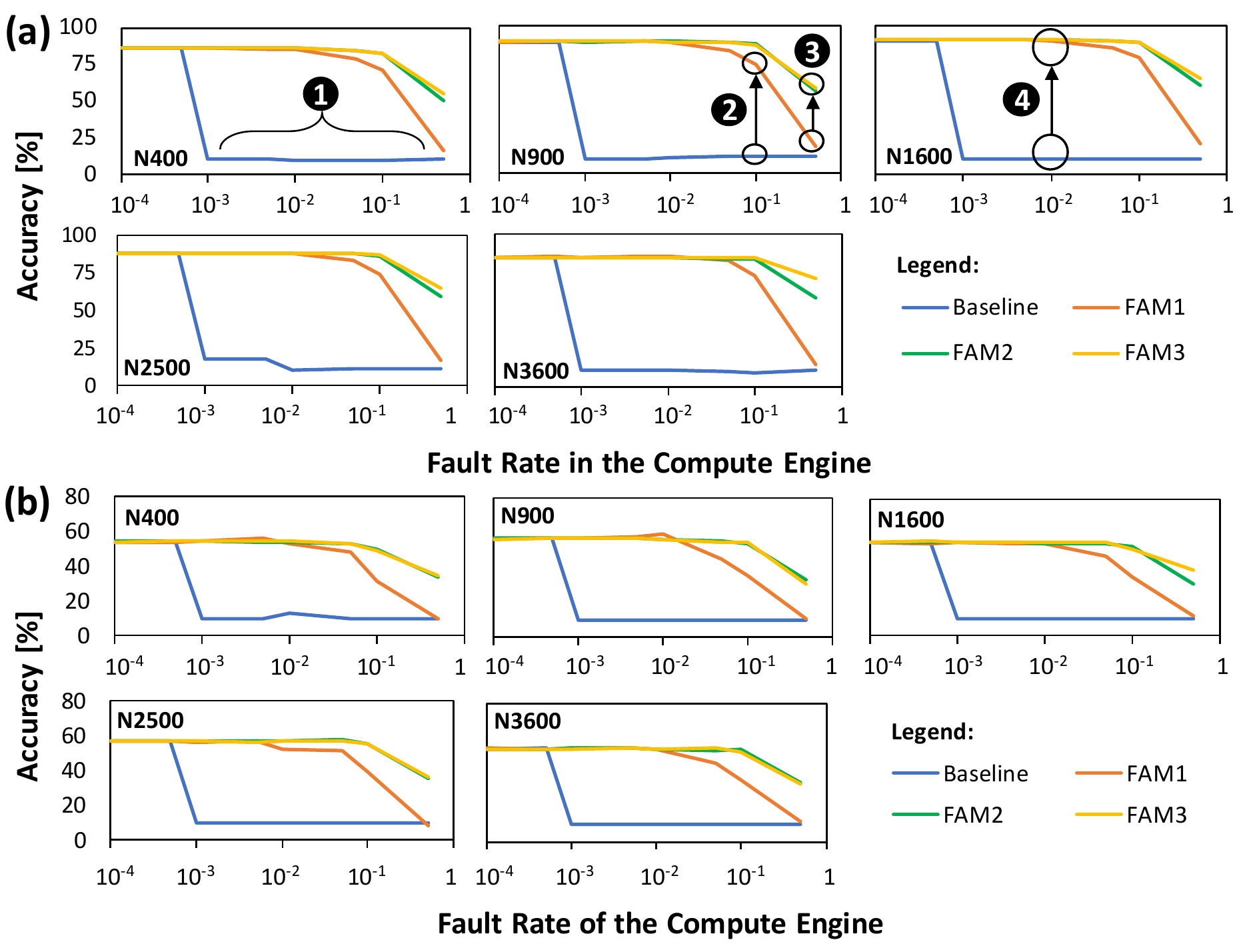}
\vspace{-0.6cm}
\caption{Accuracy profiles for different mitigation techniques (i.e., baseline, FAM1, FAM2, and FAM3), different model sizes (i.e., N400, N900, N1600, N2500, and N3600), different fault rates, and different workloads: (a) MNIST and (b) Fashion MNIST.}
\label{Fig_Results_Accuracy}
\end{figure*}
%%%% insert

%%%%%%%%%%%%%%%%%%%%%%%%
%%%%%%%%%%%%%%%%%%%%%%%%
\section{Results and Discussion}
\label{Sec_Results}

We evaluate different design aspects including accuracy, throughput, energy consumption, and area as discussed in the following.

%%%%%%%%%%%
\subsection{Maintaining Accuracy}
\label{Sec_Results_Accuracy}

Fig.~\ref{Fig_Results_Accuracy} presents the experimental results for the accuracy of different fault mitigation techniques, i.e., the baseline and our FAM-based strategies including FAM1, FAM2, and FAM3. 
We observe that the baseline suffers from a significant accuracy degradation as shown by~\circledB{1}, because it does not mitigate faults in synapses and neurons, thereby leading to unreliable SNN executions.
The significant accuracy degradation is mainly due to the fault model for faulty `$V_{mem}$ reset' operation that makes the corresponding neuron generate spikes continuously once its membrane potential $V_{mem}$ reaches the threshold potential $V_{th}$, thereby dominating the classification activity and leading to high misclassification.
We also observe that FAM1 significantly improves the SNN fault tolerance as compared to the baseline, because FAM1 avoids the use of faulty neurons, especially for faulty `$V_{mem}$ reset' operations, as shown by~\circledB{2}.
Our FAM2 improves the SNN fault tolerance even more as compared to FAM1, since FAM2 also mitigates faults in the weight registers in addition to avoiding the use of faulty neurons, as shown by~\circledB{3}.
Meanwhile, our FAM3 also significantly improves the SNN fault tolerance from baseline and FAM1, and obtains comparable accuracy to FAM2, since FAM3 mitigates faults in weight registers and selectively uses faulty neurons. 
It achieves up to 80\% accuracy improvement compared to the baseline on the MNIST dataset, as shown by~\circledB{4}. 
We also observe that the same reasons are also applicable to different workloads, thereby leading the accuracy profiles for Fashion MNIST to have similar trends to the accuracy profiles for MNIST.
These results show that \textit{our FAM strategies (FAM1, FAM2, and FAM3) are effective for mitigating permanent faults in the compute engine without retraining, across different model sizes, fault rates, and workloads}.

%%%%%%%%%%%
\subsection{Maintaining Throughput}
\label{Sec_Results_Throughput}

\begin{figure}[t]
\centering
\includegraphics[width=0.72\linewidth]{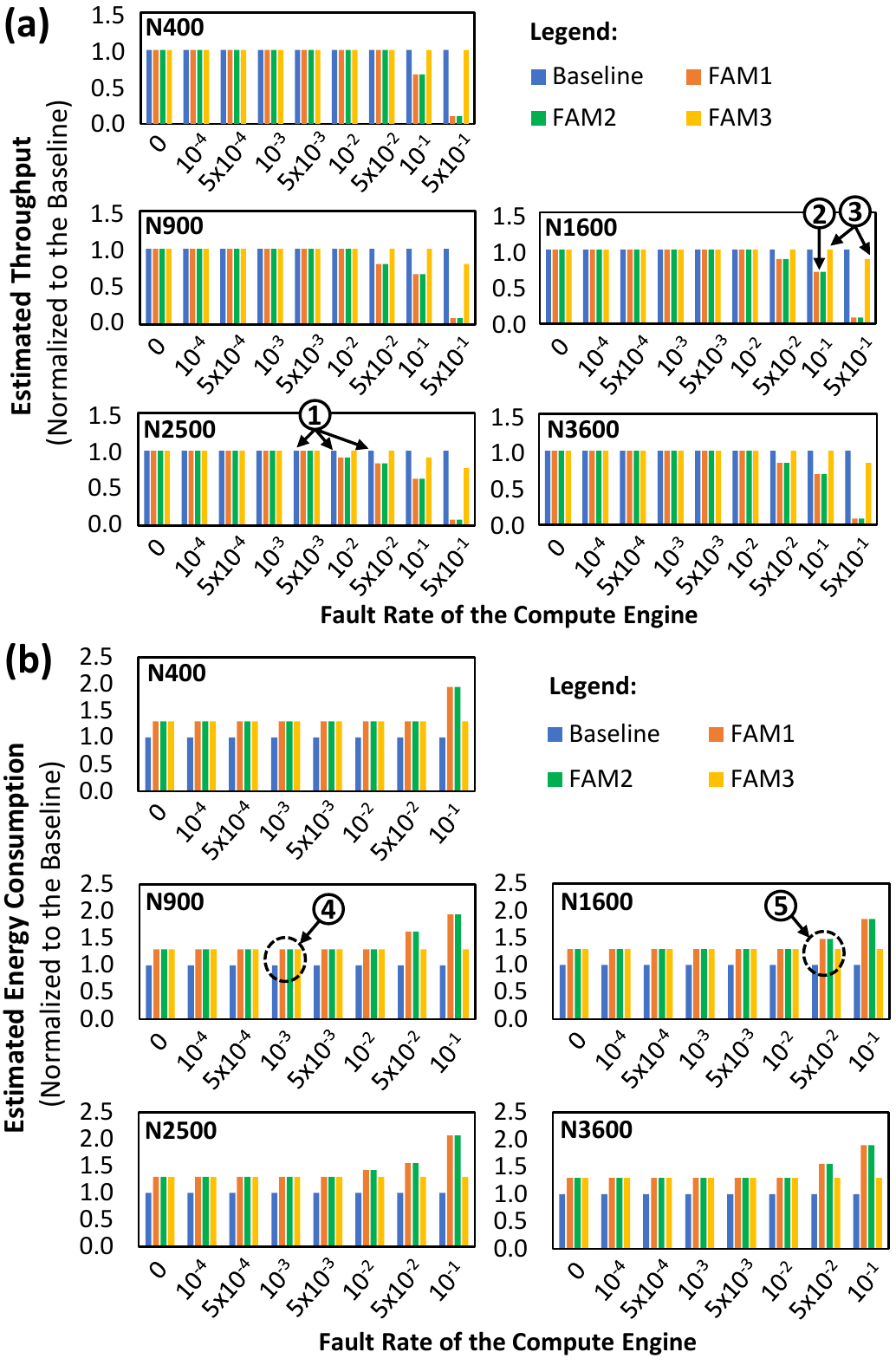}
\caption{(a) Throughput across different mitigation techniques, different model sizes, and different fault rates for both MNIST and Fashion MNIST, as they have a similar number of SNN weights and operations. (b) Energy consumption across different mitigation techniques, different model sizes, and different fault rates for both MNIST and Fashion MNIST, as they have a similar number of SNN weights and operations.}
\label{Fig_Results_ThroughputEnergy} 
\vspace{-0.4cm}
\end{figure}

Fig.~\ref{Fig_Results_ThroughputEnergy}(a) presents the experimental results on the throughput of different mitigation techniques, i.e., the baseline and our FAM strategies (FAM1, FAM2, and FAM3). 
We observe that the baseline has the highest throughput across different model sizes and fault rates, as it uses all synapses and neurons for performing SNN executions, as shown by~\circled{1}.
Meanwhile, FAM1 and FAM2 may suffer from throughput reduction because they avoid the use of faulty neurons, thereby omitting the corresponding columns of the SNN compute engine. 
For instance, FAM1 and FAM2 may suffer from 30\% throughput reduction for N1600 with 0.1 fault rate, as shown by~\circled{2}.   
Meanwhile, our FAM3 can maintain the throughput close to the baseline (e.g.,  keeping the throughput reduction below 25\% in a 0.5 fault rate), thereby improving the throughput significantly as compared to FAM1 and FAM2.
The reason is that, FAM3 omits the columns of compute engine only if the corresponding neurons have faulty `$V_{mem}$ reset' operations.
For instance, FAM3 has less than 15\% throughput reduction for N1600, as indicated by~\circled{3}.  
These results show that \textit{our FAM3 is effective for maintaining the throughput across different model sizes, fault rates, and workloads}.

%%%%%%%%%%%
\subsection{Energy Consumption and Area Overheads}
\label{Sec_Results_Overheads}

Fig.~\ref{Fig_Results_ThroughputEnergy}(b) shows the experimental results on the energy consumption of different mitigation techniques, i.e., the baseline and our FAM strategies (FAM1, FAM2, and FAM3).
We observe that different techniques have comparable energy for small fault rates, as shown by label-\circled{4}.
The reason is that small fault rates have a low probability of faulty neurons, hence the resource utilization for different techniques is similar.
For large fault rates, FAM1 and FAM2 have higher energy consumption than the baseline and FAM3, as shown by label-\circled{5}. 
The reason is that large fault rates have a high probability of faulty neurons, hence the resource utilization for different techniques is different, i.e., FAM1 and FAM2 avoid the use of faulty neurons, thereby incurring higher compute latency and energy consumption.
The baseline and FAM3 have comparable energy since FAM3 employs simple hardware enhancements: (1) multiplexing operations in each HEB which are shared for all synapses in the same column of compute engine, and (2) registers accesses in ECU, thereby minimizing the energy consumption overhead for FAM3 (i.e., within 30\%).
For area footprint, the original compute engine consumes around 6.27 $mm^2$ of area, while the one with proposed enhancements consumes around 8.56 $mm^2$ of area. 
Therefore, the proposed enhancements incur about 36.5\% of area overhead, which encompasses about 36.2\% of ECU and about 0.3\% of HEBs. 
The area of ECU dominates the total area of enhancements since it mainly employs a set of 3-bit registers (i.e., 256x256 registers), which incurs a larger area as compared to HEBs (i.e., 256x25 multiplexers).
These results show that \textit{our FAM3 achieves minimum overheads in terms of energy consumption and area across different model sizes, fault rates, and workloads}.

\smallskip
In summary, the above discussions show that \textit{our RescueSNN methodology can effectively mitigate permanent faults in the SNN chips without retraining}. 
Since our RescueSNN addresses permanent faults during both the design time and the run time, it increases the yield of SNN chips, as well as enables efficient and reliable SNN executions during their operational lifetime.
Furthermore, our RescueSNN also avoids carbon emission as it does not need any retraining, thereby offering an environment-friendly solution~\citep{Ref_Strubell_Carbon_ACL19, Ref_Strubell_Carbon_AAAI20}.    

%%%%%%%%%%%
\subsection{Further Discussion}
\label{Sec_Results_Discussion}

In general, we observe that a faulty `$V_{mem}$ reset' operation can cause significant accuracy degradation as it deteriorates the neuron from the expected behavior.
The reason is that, the generated (faulty) spikes will affect how the SNN model understands the information, since an SNN model typically employs a certain spike coding scheme, i.e., rate coding in this work. 
Therefore, a neuron with faulty `$V_{mem}$ reset' operation will generate a high number of spikes and dominate the classification activity, thereby leading to high misclassification and significant accuracy degradation. 
We also observe that, a higher number of spikes generated by faulty `$V_{mem}$ reset' operation also indicates that the SNN model performs more frequent neuron operations that correspond to spike generation. 
This condition leads to higher power/energy consumption for SNN processing, which has been observed and studied in previous works~\citep{Ref_Krithivasan_SpikeBundle_ISLPED19, Ref_Park_T2FSNN_DAC20, Ref_Putra_TopSpark_arXiv23}.

\textbf{Comparisons with Retraining Technique:} 
In a standard chip fabrication process, manufactured chips are evaluated in a wafer/chip test procedure (i.e., wafer acceptance test and chip probing test). 
This test procedure aims at evaluating the quality of each chip, including any faults in the chip~\citep{Ref_Xu_WaferWAT_Access20, Ref_Fan_FaultWaferChip_TASE22}.
In this step, the permanent faults and the corresponding fault map information from manufacturing defects are identified.
Therefore, this step does not introduce new cost, and only requires a typical cost for a standard wafer/chip test procedure~\citep{Ref_Xu_WaferWAT_Access20, Ref_Fan_FaultWaferChip_TASE22}.    
In the retraining technique, the fault map information is then incorporated in the retraining process considering how the weights and neuron operations are mapped on the SNN compute engine, i.e., so-called \textit{fault-aware training} (FAT). 
In this manner, the SNN model is expected to adapt to the presence of faults, hence maintaining high accuracy.  
This indicates that, the retraining technique requires (1) fault map information from the chip test procedure, and (2) a full training dataset, which may be unavailable due to restriction policy.
\textit{Furthermore, each chip has a unique fault map which requires its own retraining process, thereby incurring huge time and energy costs. }
Otherwise, the retraining technique will not be effective.
Meanwhile, our proposed FAM technique in RescueSNN methodology leverages the fault map information to safely map the weights and neuron operations on the SNN compute engine. 
It ensures that the SNN processing is not negatively affected by permanent faults, thereby maintaining high accuracy.
Although each chip has a unique fault map which requires a specific mapping, the cost for devising the mapping strategy is significantly lower than the cost of retraining.
Furthermore, our FAM technique does not require any training dataset, hence it is highly applicable to a wide range of SNN applications.

\textbf{Benefits and Limitations of Pruning:}
Neurons in the fully-connected (FC)-based SNN architecture shown in Fig.~\ref{Fig_SNN_PermanentFaults}(a) can be pruned while keeping the accuracy close to that of the original network, considering that a high rate of faulty `$V_{mem}$ increase' operations does not significantly degrade accuracy. 
The benefits of pruning in FC-based architecture have been demonstrated in previous work~\citep{Ref_Rathi_PruneQuantizeSNN_TCAD18}, including reduction of memory footprint and energy consumption. 
The pruning technique is suitable if we rely on offline training, i.e., an SNN model is trained offline with the training dataset, and the knowledge learnt from the training phase is kept unchanged during inference at run time. 
However, the pruning technique is not suitable if we consider SNN-based systems that need to update their knowledge regularly at run time to adapt to different operational environments (i.e., so-called \textit{dynamic environments}) such as autonomous mobile agents, e.g., unmanned ground vehicles (UGVs). 
The reason is that, SNN-based systems may encounter new input features in different environments and the offline-trained knowledge may not be representative for recognizing the corresponding classes, thereby leading to low accuracy at run time and requiring online training to update their knowledge~\citep{Ref_Putra_SpikeDyn_DAC21, Ref_Putra_lpSpikeCon_IJCNN22}.
Therefore, SNN models with unpruned neurons and unsupervised learning capabilities are beneficial for learning and recognizing new features in (unlabeled) data samples from the operational environments during online training. 
In summary, users can select which SNN model to employ depending on the design requirements. 
An alternative is employing the FC-based SNNs shown in Fig.~1(a) with/without pruning since they can enable multiple benefits, such as high accuracy when employing STDP-based learning under unsupervised settings, and efficient online training capabilities.

%%%%%%%%%%%%%%%%%%%%%%%%
%%%%%%%%%%%%%%%%%%%%%%%%
\section{Conclusion}
\label{Sec_Conclusion}

We propose RescueSNN, a novel methodology for mitigating permanent faults in SNN chips. 
RescueSNN leverages the fault map of compute engine to perform fault-aware mapping for SNN weights and operations, and employs efficient hardware enhancements for the proposed mapping technique. 
The results show that RescueSNN improves the SNN fault tolerance without retraining. 
As a result, faulty SNN chips can be rescued and used for reliable SNN processing during their operational lifetime. 

%%%%%%%%%%%%%%%%%%%%%%%%%%%%%%%%%%%%%%%%%%%%%%%%%%%%%%%
%%%%%%%%%%%%%%%%%%%%%%%%%%%%%%%%%%%%%%%%%%%%%%%%%%%%%%%
\section*{Data Availability Statement}
\label{Sec_DataStatement}

Publicly available datasets were used in this study. 
These datasets can be found at the following sites: http://yann.lecun.com/exdb/mnist/ (MNIST),
http://fashion-mnist.s3-website.eu-central-1.amazonaws.com/ (Fashion MNIST).

%%%%%%%%%%%%%%%%%%%%%%%%%%%%%%%%%%%%%%%%%%%%%%%%%%%%%%%
%%%%%%%%%%%%%%%%%%%%%%%%%%%%%%%%%%%%%%%%%%%%%%%%%%%%%%%
\section*{Funding}
\label{Sec_Funding}

This work was supported in parts by the Center for Artificial Intelligence and Robotics (CAIR), funded by Tamkeen under the NYUAD Research Institute Award CG010, and the Center for Cyber Security (CCS), funded by Tamkeen under the NYUAD Research Institute Award G1104. This work was also partially supported by the project ``eDLAuto: An Automated Framework for Energy-Efficient Embedded Deep Learning in Autonomous Systems'', funded by the NYUAD Research Enhancement Fund (REF). The authors acknowledge TU Wien Bibliothek for the publication fee support through its Open Access Funding Program.

%%%%%%%%%%%%%%%%%%%%%%%%%%%%%%%%%%%%%%%%%%%%%%%%%%%%%%%
%%%%%%%%%%%%%%%%%%%%%%%%%%%%%%%%%%%%%%%%%%%%%%%%%%%%%%%
% \section*{Supplemental Data}

%%%%%%%%%%%%%%%%%%%%%%%%%%%%%%%%%%%%%%%%%%%%%%%%%%%%%%%
%%%%%%%%%%%%%%%%%%%%%%%%%%%%%%%%%%%%%%%%%%%%%%%%%%%%%%%

% \bibliographystyle{Frontiers-Harvard}
\bibliographystyle{frontiersinSCNS_ENG_HUMS} % for Science, Engineering and Humanities and Social Sciences articles, for Humanities and Social Sciences articles please include page numbers in the in-text citations
\bibliography{Bibliography}

%%% Make sure to upload the bib file along with the tex file and PDF
%%% Please see the test.bib file for some examples of references
\end{document}